\documentclass[final, twocolumn,5p,times, authoryear]{elsarticle}
\usepackage{amsmath}
\usepackage{booktabs}
\usepackage{url}
\usepackage{algorithm}
\usepackage{amsmath}
\usepackage{algpseudocode}
\usepackage{amsmath}
\usepackage{array}
\usepackage[colorlinks=False]{hyperref}
\usepackage[round]{natbib}
\usepackage{lineno}
\usepackage{multirow}

\usepackage{amssymb}

\usepackage{orcidlink}
\begin{document}
\begin{frontmatter}



\title{Seamless High-Resolution Terrain Reconstruction: A Prior-Based Vision Transformer Approach}


\author[inst1]{Osher Rafaeli\corref{cor1}\raisebox{0.5ex}{\orcidlink{0000-0002-7097-7568}}}
\ead{osherr@post.bgu.ac.il}
\cortext[cor1]{Corresponding author.}

\author[inst1,inst2]{Tal Svoray\raisebox{0.5ex}{\orcidlink{0000-0003-2243-8532}}}

\author[inst1]{Ariel Nahlieli\raisebox{0.5ex}{\orcidlink{0009-0001-0633-1842}}}

\affiliation[inst1]{
    organization={Department of Environmental, Geoinformatics and Urban Planning Sciences, 
    Ben-Gurion University of the Negev},
    country={Israel}
}

\affiliation[inst2]{
    organization={Department of Psychology, 
    Ben-Gurion University of the Negev},
    country={Israel}
}

\begin{abstract}
High-resolution elevation data is essential for hydrological modeling, hazard assessment, and environmental monitoring; however, globally consistent, fine-scale Digital Elevation Models (DEMs) remain unavailable. Very high-resolution single-view imagery enables the extraction of topographic information at the pixel level, allowing the reconstruction of fine terrain details over large spatial extents. In this paper, we present single-view-based DEM reconstruction shown to support practical analysis in GIS environments across multiple sub-national jurisdictions. Specifically, we produce high-resolution DEMs for large-scale basins, representing a substantial improvement over the 30~m resolution of globally available Shuttle Radar Topography Mission (SRTM) data. The DEMs are generated using a prior-based monocular depth foundation (MDE) model, extended in this work to the remote sensing height domain for high-resolution, globally consistent elevation reconstruction. We fine-tune the model by integrating low-resolution SRTM data as a global prior with high-resolution RGB imagery from the National Agriculture Imagery Program (NAIP), producing DEMs with near LiDAR-level accuracy. Our method achieves a $100\times$ resolution enhancement (from 30~m to 30~cm), exceeding existing super-resolution approaches by an order of magnitude. Across two diverse landscapes, the model generalizes robustly, resolving fine-scale terrain features with a mean absolute error of less than 5~m relative to LiDAR and improving upon SRTM by up to 18\%. Hydrological analyses at both catchment and hillslope scales confirm the method’s utility for hazard assessment and environmental monitoring, demonstrating improved streamflow representation and catchment delineation. Finally, we demonstrate the scalability of the framework by applying it across large geographic regions.
\end{abstract}


\end{frontmatter}


\section{Introduction}

High-resolution topography is fundamental to understanding Earth surface processes, with applications ranging from hydrological and ecological modeling to urban planning, navigation systems, and forest management \citep{WILSON2012107, fernandez2016analysis, khan2024terrestrial}. Digital elevation models (DEMs) provide a three-dimensional representation of the terrain surface, typically stored as a grid, and constitute a core dataset for such analyses \citep{Yang20072011}. 

DEMs are commonly derived from remotely sensed data, including synthetic aperture radar (SAR), airborne light detection and ranging (LiDAR), and RGB photogrammetry \citep{guth2021digital}. These sources enable DEM generation across a wide range of spatial scales, from highly detailed local topography \citep{schumann2008comparison} to near-global coverage \citep{BERRY200717}. At the global scale, DEMs are most frequently derived from the Shuttle Radar Topography Mission (SRTM), which provides near-worldwide coverage at a spatial resolution of 30~m \citep{rodriguez2006global}. 

Despite their widespread use, environmental studies are often constrained by the limited availability of high-resolution DEMs. This limitation primarily arises from the high cost and restricted spatial coverage of airborne LiDAR surveys \citep{BHARDWAJ2016125}. While photogrammetric DEMs offer a more accessible alternative, they require overlapping image pairs and are susceptible to stereo-matching errors, which can introduce gaps or distortions \citep{colomina2014unmanned}. Furthermore, high-resolution DEMs are infrequently updated, limiting their applicability for dynamic analyses at plot-, hillslope-, and catchment scales \citep{wang2021challenges}.

To address the growing demand for high-resolution DEMs (HR-DEMs), super-resolution (SR) approaches have been developed to enhance the spatial resolution of existing DEMs using machine learning and interpolation techniques \citep{s22030745}. Deep learning-based methods, including SRCNNs, ESRGANs, and transformer-based models, have demonstrated improvements over traditional interpolation \citep{10418045}. However, SR approaches remain constrained by fixed upscaling factors (typically $4\times$--$16\times$), limiting their ability to reconstruct arbitrarily high-resolution terrain \citep{YAO20241, isprs-annals-X-2-2024-185-2024, Wang31122024}.

In parallel, advances in computer vision have driven rapid progress in monocular depth estimation (MDE), which infers depth from a single RGB image \citep{MING202114}. Unlike SR, which enhances existing elevation data, MDE reconstructs scene geometry directly from visual cues such as texture, structure, and object semantics \citep{Zhan_2018_CVPR, Hermann_2020}. While MDE can produce high-resolution depth maps aligned with image resolution, its predictions are inherently relative, lacking absolute scale and global consistency, which limits its applicability for DEM reconstruction \citep{Florea_2025, hong2025depth2elevation, cambrin2024depth}.

To address the growing demand for high-resolution DEMs (HR-DEMs), super-resolution (SR) approaches have been developed to enhance the spatial resolution of existing DEMs using machine learning (ML) and spatial interpolation techniques \citep{s22030745}. Deep learning-based methods, including super-resolution convolutional neural networks (SRCNNs), generative adversarial networks (ESRGANs), and vision transformer-based models such as TTSR, have demonstrated substantial improvements over traditional interpolation methods \citep{10418045}. More recent studies have further explored the integration of auxiliary remote sensing modalities, such as shaded relief \citep{HUANG2024104014}, hydrological constraints \citep{cao2025integrating}, and relative depth information \citep{huang2025multimodal, ahmed6105125geofusion}, to improve DEM super-resolution performance. However, SR approaches remain constrained by fixed upscaling factors, typically limited to $4\times$--$16\times$, and their output resolution is largely determined by the model architecture and training data rather than being adaptively driven by the input \citep{YAO20241, isprs-annals-X-2-2024-185-2024, Wang31122024}.

Monocular height estimation (MHE) is increasingly adopting monocular depth estimation (MDE) foundation models \citep{SONG2026155}. MHE approaches based on foundation models improve prediction accuracy and generalization; however, they primarily focus on object-level height estimation, producing normalized digital surface models (nDSMs) for buildings and vegetation at a local, patch-based scale, rather than enabling continuous terrain reconstruction \citep{cambrin2024depth, chen2023htcdcnetmonocularheight, hong2025depth2elevation, TOLAN2024113888}. Consequently, these methods are not well suited for reconstructing full surface topography in open environments or for downstream applications such as hydrological analysis.

Here, we employ a foundation model-based framework that extends the prior-based monocular paradigm to DEM reconstruction, enabling the integration of globally available elevation priors with high-resolution optical imagery. This approach addresses two key limitations: (1) overcoming the scale constraints of SR methods by enabling resolution enhancement of up to $100\times$ without predefined scaling factors, and (2) incorporating global elevation context into MDE to ensure spatially continuous and metrically accurate predictions.

To achieve this, we define three objectives: (1) to comprehensively validate prior-based high-resolution MDE using coarse, globally available elevation data; (2) to evaluate model performance across diverse landscape types, including vegetated and bare terrains, using airborne LiDAR-derived DEMs as reference; and (3) to demonstrate the capability of the framework to generate seamless, high-resolution DEMs with global context over large geographic regions, with a focus on the United States. 

The main contributions of this study are:

\begin{itemize}
  \item We provide the first large-scale evaluation of an elevation-prior monocular foundation model in natural environments, encompassing both vegetated and bare landscapes.

  \item We integrate freely available SRTM DEMs as global elevation priors, enabling metrically consistent and geographically aligned height estimation.

  \item We develop a patch-wise reconstruction framework that produces seamless, high-resolution DEMs suitable for downstream terrain analysis, including slope, aspect, stream network extraction, and basin delineation.
\end{itemize}

\begin{figure*}[ht!]
    \centering
    \includegraphics[width=1\linewidth]{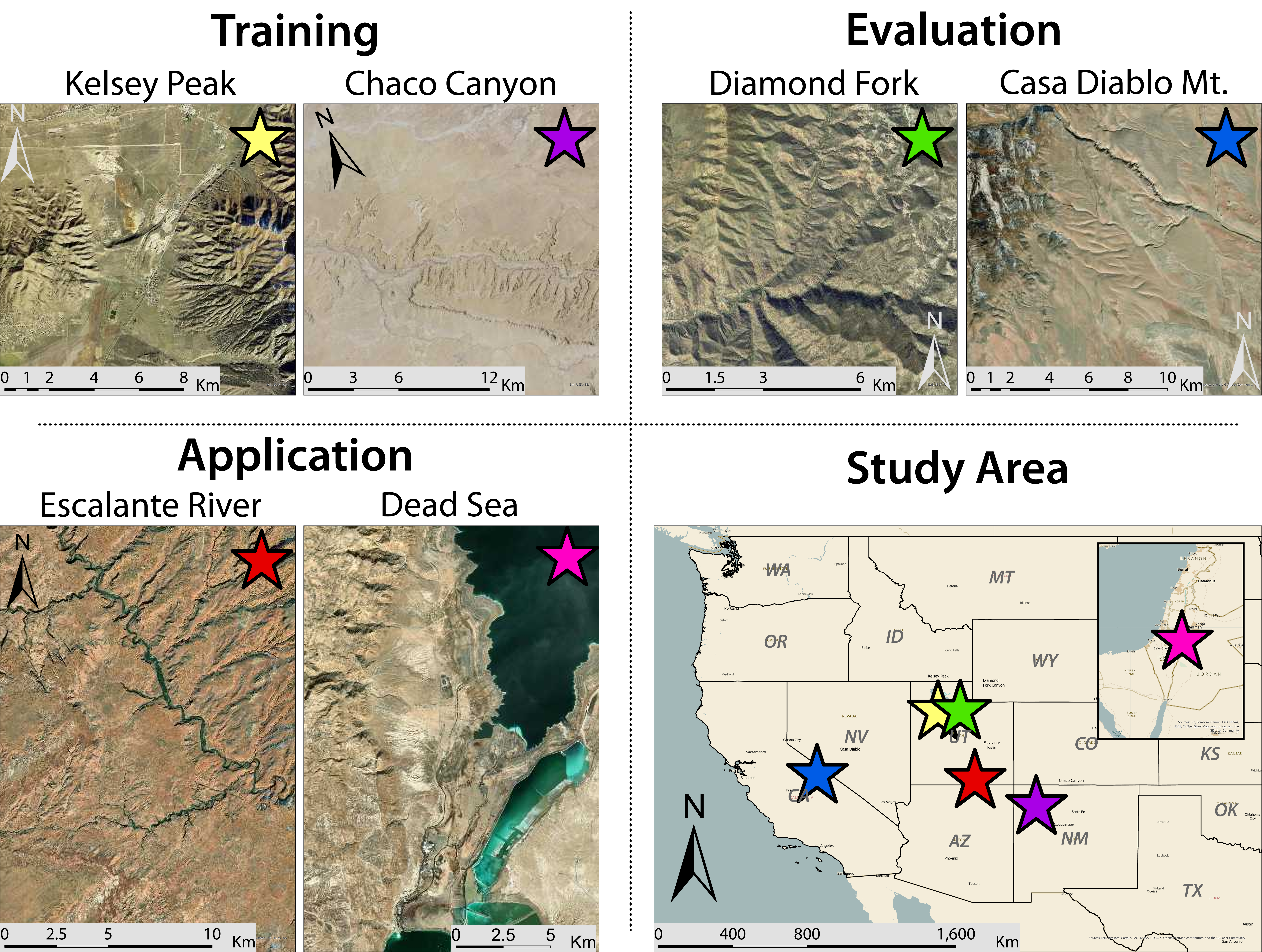}
    \caption{Our study sites for training and evaluation are located in four different states in the United States. For natural vegetated areas, we used Kelsey Peak and Diamond Fork Canyon in Utah. For natural, bare sites, we used Chaco Canyon in New Mexico and Casa Diablo Mountain in California. For model application, we used the Escalante River in Utah, and the Dead Sea in Israel.}
    \textit{}
    \label{study_area}
\end{figure*}

\begin{table*}[ht!]
\centering
\caption{Summary of the datasets used for elevation estimation across six study sites.}
\begin{tabular}{lccccc}
\hline
\textbf{Site} & \textbf{Landscape Type} & \textbf{RGB (NAIP)} & \textbf{LiDAR DEM} & \textbf{SRTM DEM} & \textbf{Area (km$^2$)} \\
\hline
Kelsey Peak, UT             & Vegetated  & 60 cm/px & 50 cm/px & 30 m/px & 22 \\
Diamond Fork Canyon, UT     & Vegetated  & 60 cm/px & 50 cm/px & 30 m/px & 22 \\
Chaco Canyon, NM            & Bare       & 60 cm/px & 50 cm/px & 30 m/px & 70 \\
Casa Diablo Mountain, CA    & Bare       & 60 cm/px & 1 m/px & 30 m/px & 40 \\
\hline
\label{tab:datasets}
\end{tabular}
\end{table*}

\begin{table*}[ht!]
\centering
\caption{Statistical metrics of elevation distribution for DEMs across six study sites. Slope statistics are shown in parentheses with units in degrees (°), while elevation units are in meters (m).}
\begin{tabular}{lccccc}
\hline
\textbf{Site Elevation} & \textbf{Mean (m)} & \textbf{Std (m)} & \textbf{Skewness} & \textbf{Kurtosis} & \textbf{Bimodality Coefficient} \\
\hline
Kelsey Peak, UT             & 1863.03 (61.32°) & 194.32 (13.23°)  & 0.55 (-1.30) & 2.25 (4.40) & 0.581 (0.612) \\
Diamond Fork Canyon, UT     & 1950.41 (59.73°) & 228.44 (17.46°)  & 0.00 (-1.02) & 2.11 (3.44) & 0.475 (0.592) \\
Chaco Canyon, NM            & 1943.40 (19.92°) & 46.47 (18.67°)   & 1.37 (1.51)  & 4.95 (4.67) & 0.579 (0.703) \\
Casa Diablo Mountain, CA    & 1701.39 (32.23°) & 139.57 (16.27°)  & 0.30 (0.88)  & 2.27 (3.33) & 0.481 (0.532) \\
\hline
\label{tab:stat}
\end{tabular}
\end{table*}

\begin{figure}[ht!]
    \centering
    \includegraphics[width=1\linewidth]{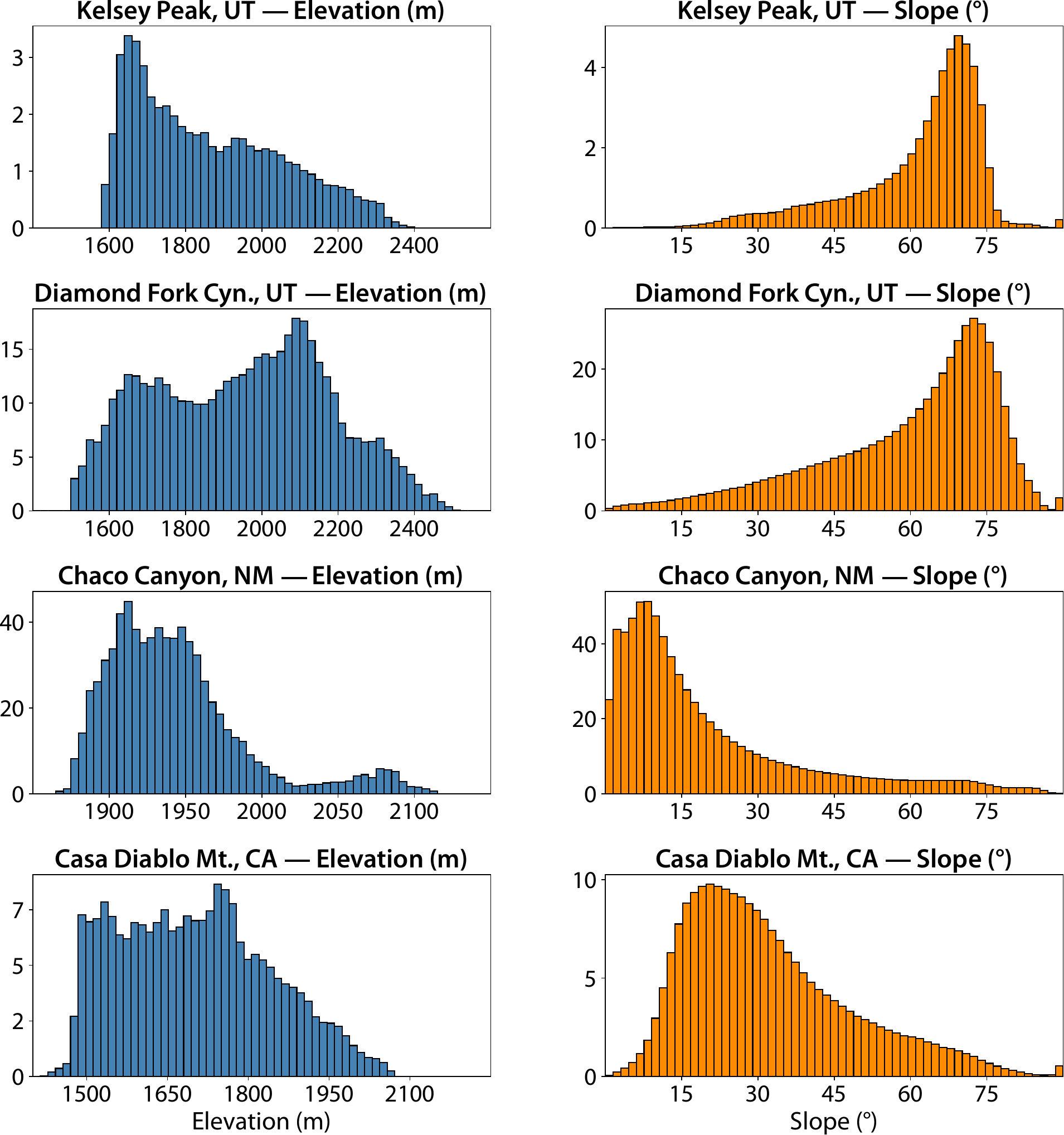}
    \caption{Histograms of elevation and slope across the four study sites indicate that naturally vegetated regions generally have steeper slopes compared to bare terrain.}
    \phantomsection
    \label{fig:dsm_elevation_slope_histograms}
\end{figure}

\section{Materials and Methodology}

\begin{figure}[ht!]
    \centering
    \includegraphics[width=1\linewidth]{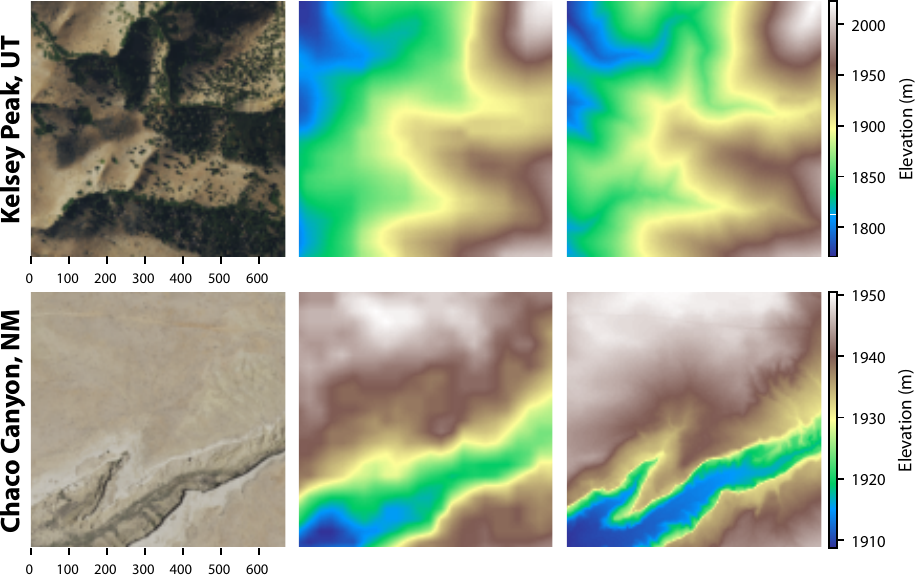}
    \caption{Sample patches from the training dataset include NAIP RGB imagery, SRTM data as priors, and high-resolution LiDAR DEMs as ground truth.}
    \label{austin_kellum_chaco}
\end{figure}

\subsection{Study area and dataset}

To examine the proposed framework, we agglomerated a dataset from publicly available sources. This dataset was used for model training, testing and generalization. It includes: (1) HR-RGB imagery from the National Agriculture Imagery Program (NAIP) \citep{usgs_naip}; (2) global 30-m SRTM LR elevation data \citep{srtm2013global}; and (3) HR LiDAR-derived DEMs obtained from the OpenTopography platform \citep{opentopography}. The dataset covers several environments in the US, and is consistent across two representative open landscapes: vegetated- and bare-terrain. For each type, two sites were selected: one for training and the other for testing (Figure \ref{study_area} and Table \ref{tab:datasets}).

The vegetated landscape is represented by Kelsey Peak (40.4580°,-112.3644°) \citep{lakebonneville2025} and Diamond Fork Canyon (40.1047°, -111.4355°), both in Utah \citep{jones2018diamondfork}. The bare, or sparsely vegetated, landscape is  represented by Chaco Canyon in northwestern New Mexico (36.0544°,-107.9470°) \citep{dorshow2019chaco} and by the Casa Diablo Mountain zone in the Californian desert (37.5919°,-118.5232°) \citep{bishop2025tuff}.

\begin{figure*}[ht!]
    \centering
    \includegraphics[width=1\linewidth]{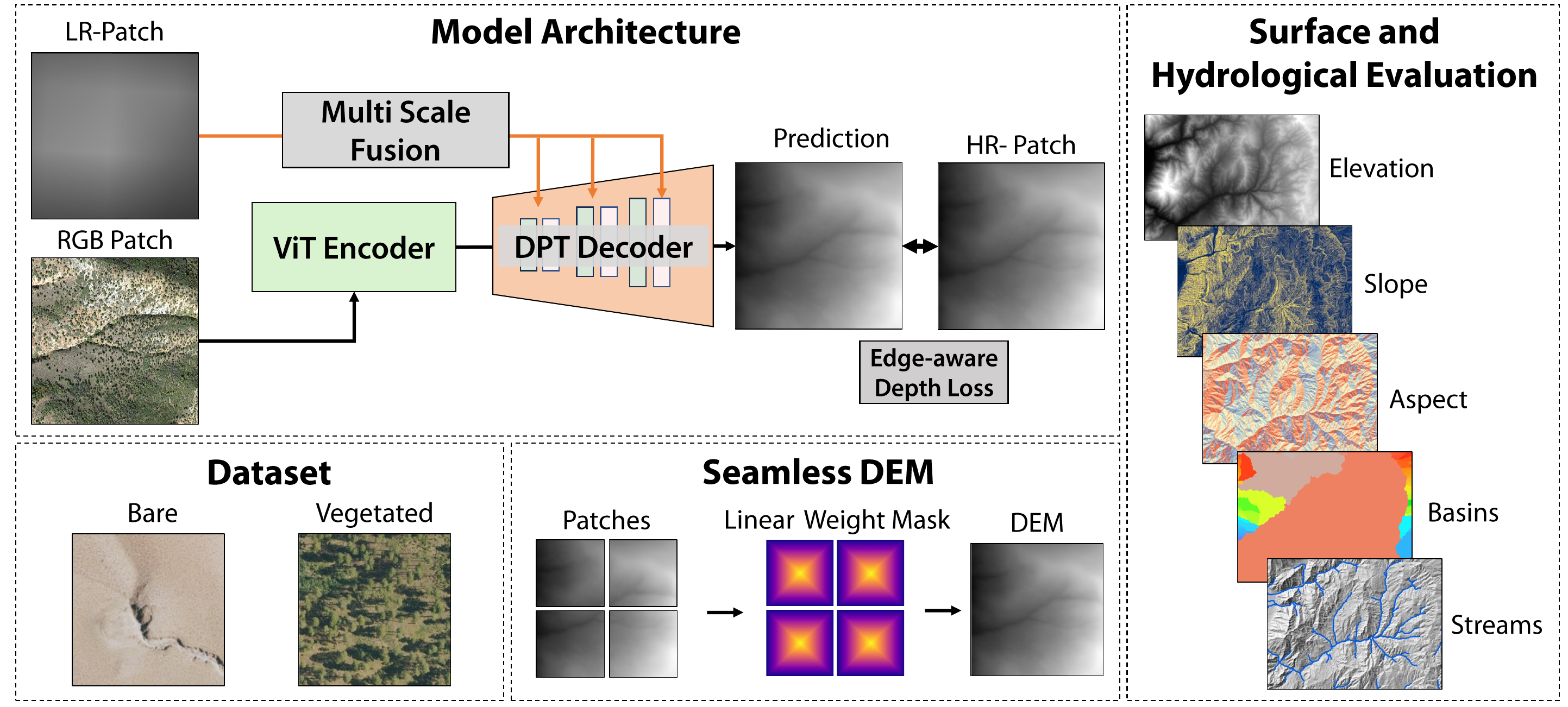}
    \caption{Proposed Framework: Dataset and  Landscape categories: The model was trained and evaluated on natural vegetated and natural bare lands. All model components were fine-tuned using NAIP 3-channel (RGB) aerial images with LR-SRTM and HR-DEM extracted from HR LiDAR-based DEM. This mitigates edge artifacts and ensures smooth transitions between overlapping patches. We applied a linear weighted mask.}
    \textit{}
    \label{framework}
\end{figure*}

The two landscape types were characterized by large variation in elevation range. The natural vegetated areas were steeper, with Diamond Fork Canyon and Kelsey Peak in Utah showing high mean elevation values and large standard deviations (194–228 m) indicating irregular mountainous terrain. The value of the mean hillslope gradient in Diamond Fork Canyon was 59.73$^\circ$, and in Kelsey Peak, it was 61.32$^\circ$. Although these values are very high, the low standard deviation  ($<20^\circ$) indicates a relatively uniform terrain.

Both Chaco Canyon, NM, and Casa Diablo Mountain, CA, exhibited lower variability in elevation values than the vegetated sites in Utah. In particular, skewness and kurtosis of Chaco Canyon were higher due to the asymmetry between the steep canyon and the plateau. The mean hillslope gradient was 19.92$^\circ$, with a bimodal distribution resulting from the contrast between the broad plateaus and the steep canyons. A smoother distribution of elevation values was obtained for Casa Diablo, which features more uniform terrain and broad hillslopes, with a mean hillslope gradient of 32.23$^\circ$. Standard deviation was higher than in the vegetated sites due to steep narrow canyons and stream incisions (Figure \ref{fig:dsm_elevation_slope_histograms} and Table \ref{tab:stat}).

All elevation and NAIP datasets were geo-referenced and co-registered according to the North American Datum (NAD) of 1983 UTM coordinate system \citep{fgdc1998metadata}. Each landscape type was divided into two comparable subsets, with one site used for training and the other for evaluation. Kelsey Peak and Chaco Canyon were specifically selected for fine-tuning. This selection was intended to create a balanced train–test split and to expose the model to diverse terrain conditions, thereby supporting a  comprehensive evaluation and reducing overfitting likelihood to any landscape type.

The generalization phase (application to another site) was conducted over two wide regions,  with the objective to generate  HR-DEMs. The two additional sites are: the Escalante River watershed, a tributary of the Colorado River, in Utah (\textasciitilde 474.69\,km$^2$), with standard 60-cm NAIP imagery; and the Dead Sea shore in Israel, where SRTM and high-resolution 15-cm orthomosaic data were available since 2023. The Dead Sea region is known for salt-induced sinkholes \citep{SEVIL2023108732}.

In addition to the main experiment on SRTM-LiDAR dataset, the proposed framework was compared with DEM super-resolution (DEMSR) methods. To that end, a recently introduced dataset, the DEM-OPT-Depth SR Dataset, was used. This publicly available test set contains 5-m RGB, 30-m LRDEM, and 7.5-m HRDEM data. We fine-tuned the data using only 33\% (231 samples) and tested it on the remaining 67\% (500 samples) \citep{huang2025multimodal}. We report here both metrics from the original MFSR paper, and the results of our model’s performance on the DEM-OPT-Depth SR Dataset.

\subsection{Model architecture}

Current state-of-the-art monocular depth foundation models \citep{yang2024depth, birkl2023midas} generally follow architectures similar to the Dense Prediction Transformer (DPT) framework \citep{ranftl2021vision}. These  typically employ a DINO self-supervised encoders to extract hierarchical image representations across multiple stages. The input image is first tokenized into patches, and the transformer processes these tokens to produce multi-scale feature maps that capture global context and local structural information \cite{oquab2024dinov2learningrobustvisual}.

\textbf{ViT Encoder:} We used a pretrained ViT encoder from the Depth Anything V2 model, which builds on the largest DINOv2 and leverages its inherent generalization ability \citep{yang2024depthv2}. In particular, Depth Anything V2 introduces several key training advantages: (1) It overcomes  manual image annotation limitations by incorporating synthetically annotated data during training, with fine-detailed labels; (2) It employs a teacher–student architecture to obtain a lightweight, yet accurate, model. The model is initially trained on synthetic images, after which the teacher model generates pseudo masks on unannotated real images. Finally, a smaller student model is trained on these pseudo-annotated images; and (3) Its annotation pipeline integrates SAM for image sampling, utilizing a combination of real and pseudo-labeled depth maps derived from 62 million images \citep{yang2024depthanythingunleashingpower}. 

Depth Anything V2 is benchmarked on standard datasets such as KITTI, Sintel, and others, and it outperforms previous foundation MDE models, including MiDaS v3.1 from Intel Labs \citep{birkl2023midasv31model}. Its superior performance is further demonstrated on the DA-2K dataset, which includes eight different image categories, 9\% of which consist of aerial and fine-detail imagery.

\textbf{DPT Decoder:} The DPT decoder reassembles features from different stages into image-like representations using the reassemble operation. Finally, a sequence of convolutional blending steps is applied to merge features across stages and create a dense depth map, to produce full resolution predictions using a convolutional decoder. For monocular depth estimation, we observe an improvement of up to 28\% in relative performance when compared with a state-of-the-art fully-convolutional network \citep{ranftl2021vision}. 

\textbf{Prior Fusion Block:} PromptDA proposes a concise prior-fusion architecture that is tailored for a DINOv2 encoder, a DPT-based decoder, and a prior injection mechanism to enable  integration of low-resolution (LR) depth information \citep{lin2025promptingdepth4kresolution}. priors are bilinearly resized to match the current scale of the spatial dimensions $R \in \mathbb{R}^{1 \times H_i \times W_i}$. The resized depth map is then passed through a shallow convolutional network to extract depth features. These are projected to match the channel dimensions of image features $F_i \in \mathbb{R}^{C_i \times H_i \times W_i}$ using a zero-initialized convolutional layer. Subsequently, depth features are added to intermediate features of the DPT decoder for depth prediction.

\subsection{Fine Tuning}

Typically, tuning a prior-based monocular model requires LR depth information as priors and precise ground truth (GT) depth for backpropagation. Here, because the data are transitioned to elevation values, we trained the model using: (1) global DEM 30-m SRTM; and (2) HR LiDAR-derived DEMs (50\,cm and 1\,m). For accurate predictions, we tuned all weights during the initial training. This training loop, was specifically designed for surface elevation estimation, was conducted for 10 epochs. A samples of training patches is shown in Fig. \ref{austin_kellum_chaco}.

\subsection{Seamless DEM reconstruction}

Since each patch is predicted separately due to computer resource limitations, directly stitching the patches together can create visible discontinuities between patches, which may negatively affect downstream DEM analyses such as flow accumulation and stream network extraction. To ensure smooth transitions between patches, we implemented a distance-based linear blending method for overlapping regions \citep{6619010}. This technique assigns higher weights to image pixels that are farther from the patch edges and lower weights to pixels near the edges of the patch. Overlapping areas are then combined using spatially varying weights, ensuring a seamless mosaic and mitigating boundary artifacts. 

Specifically, for each patch, a 2D linear mask is constructed by computing the minimum normalized distance from each pixel to the patch boundaries. This mask increases smoothly from 0 at the patch boundaries to 1 at its center, promoting gradual transitions in overlapping areas. The final blended value at each pixel is obtained as a weighted average of all overlapping patches using the corresponding weights.

\subsection{Surface and Hydrological Evaluation}
Two commonly used elevation error evaluation metrics, the mean absolute error (MAE) and the root mean square error (RMSE), were used (Eqs.~\ref{eq:mae} and \ref{eq:rmse}):

\begin{equation}
\text{MAE} = \frac{1}{N} \sum_{i=1}^{N} \left| y_i - \hat{y}_i \right|,
\label{eq:mae}
\end{equation}

\begin{equation}
\text{RMSE} = \sqrt{ \frac{1}{N} \sum_{i=1}^{N} (y_i - \hat{y}_i)^2 },
\label{eq:rmse}
\end{equation}
where $y_i$ denotes the pixel value of the original DEM, $\hat{y}_i$ denotes the corresponding pixel value of the filled DEM, and $N$ is the number of void pixels.

Since the hillslope gradient is closely related to relative elevation accuracy, we also calculated RMSE for similarity between topographic slope and aspect values (Eqs.~\ref{eq:slope} and \ref{eq:aspect}):

\begin{equation}
\text{Slope} = \sqrt{ \left( \frac{\partial z}{\partial x} \right)^2 + \left( \frac{\partial z}{\partial y} \right)^2 },
\label{eq:slope}
\end{equation}

\begin{equation}
\text{Aspect} = \tan^{-1}\left( \frac{ \frac{\partial z}{\partial y} }{ \frac{\partial z}{\partial x} } \right),
\label{eq:aspect}
\end{equation}
where $\frac{\partial z}{\partial x}$ and $\frac{\partial z}{\partial y}$ represent the partial derivatives of the elevation values in the $x$ and $y$ directions, respectively.

To assess the relevance of the predicted DEMs to hydrological applications \citep{https://doi.org/10.1029/2021WR030998}, we extracted stream networks using D8, the Deterministic 8-direction flow model, \citep{ocallaghan1984drainage, Svoray01092004}, which included the Fill, Flow Direction, and Flow Accumulation tools in ArcGIS Pro. The resulting stream maps were compared with ground truth using common binary segmentation metrics, i.e., Intersection over Union (IoU), precision, recall, F1-score, and overall accuracy, as defined below:

\begin{equation}
\text{IoU} = \frac{|A \cap B|}{|A \cup B|}
\label{eq:iou}
\end{equation}

\noindent
\begin{minipage}{0.48\textwidth}
\begin{equation}
\text{Precision} = \frac{TP}{TP + FP}
\label{eq:precision}
\end{equation}
\end{minipage}
\hfill
\begin{minipage}{0.48\textwidth}
\begin{equation}
\text{Recall} = \frac{TP}{TP + FN}
\label{eq:recall}
\end{equation}
\end{minipage}
\begin{equation}
\text{F1 Score} = 2 \times \frac{\text{Precision} \times \text{Recall}}{\text{Precision} + \text{Recall}}
\label{eq:f1_score}
\end{equation}

\begin{equation}
\text{Accuracy} = \frac{TP + TN}{TP + TN + FP + FN},
\label{eq:accuracy}
\end{equation}
where TP, TN, FP, and FN denote, respectively,  true positive, true negative, false positive, and false negative values, and \(A\) and \(B\) are predicted and ground truth segments, respectively. 

To further evaluate the hydrological consistency of the predicted DEMs, we also analyzed the contributing areas of sub-basins. Specifically, for each study area, ten representative sub-basins were delineated based on the flow accumulation outputs. The contributing area of each sub-basin was then computed and compared against the corresponding ground truth delineations. Similar to the stream network evaluation, we quantified the agreement between predicted and reference sub-basin extents using the same binary segmentation metrics (IoU, precision, recall, F1-score), treating the sub-basin areas as binary masks.

\subsection{Technical details}

During the training process, the batch size was set to two using a single NVIDIA RTX A4000 GPU. We used the Adam optimizer with a learning rate of 5e-5. The loss function was a composite of the standard L1 loss and an edge-aware term to supervise both the predicted elevation ($\hat{E}$) and its spatial gradients. The edge loss weight $\lambda$ was set to 0.5 The overall loss function is defined as:

\begin{equation}
\mathcal{L}_{\text{edge}} = \mathcal{L}_1(E_{\text{gt}}, \hat{E}) + \lambda \cdot \mathcal{L}_{\text{grad}}(E_{\text{gt}}, \hat{E}),
\end{equation}

where the gradient loss term is calculated as:

\begin{equation}
\mathcal{L}_{\text{grad}}(E_{\text{gt}}, \hat{E}) = 
\left| \frac{\partial (\hat{E} - E_{\text{gt}})}{\partial x} \right| + 
\left| \frac{\partial (\hat{E} - E_{\text{gt}})}{\partial y} \right|.
\end{equation}

\section{Results}

\begin{figure}[ht!]
    \centering
    \includegraphics[width=1\linewidth]{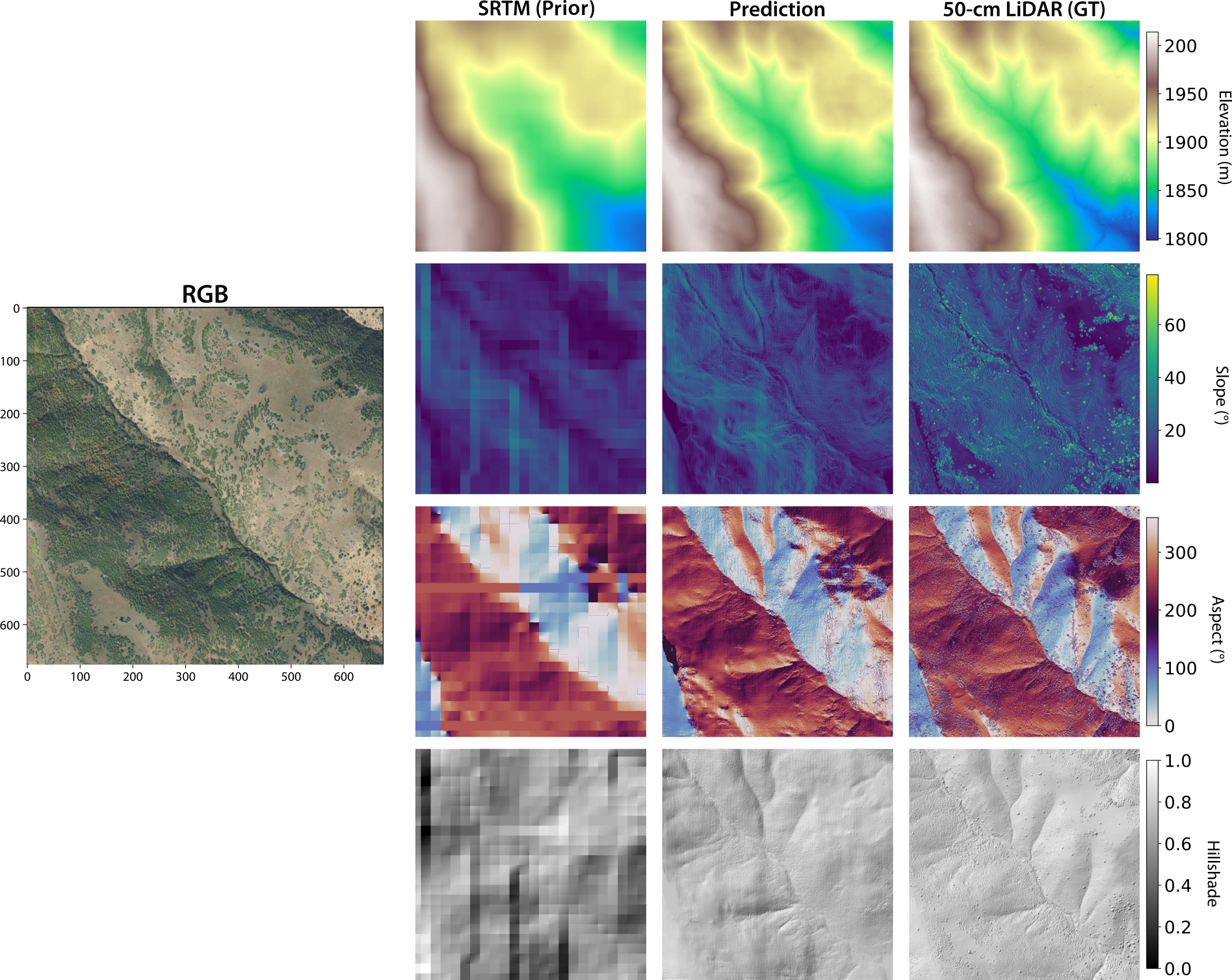}
    \caption{Visual Evaluation - DEM estimation in natural vegetated landscape, elevation in meters, slope in degrees, aspect in radial direction, and hillshade visualization. The global SRTM general terrain was preserved for elevation prediction, while RGB information added detailed terrain features, such as small streams that are clearly visible in the slope, aspect, and hillshade maps.}
    \label{veg_low}
\end{figure}

\begin{figure}[ht!]
    \centering
    \includegraphics[width=1\linewidth]{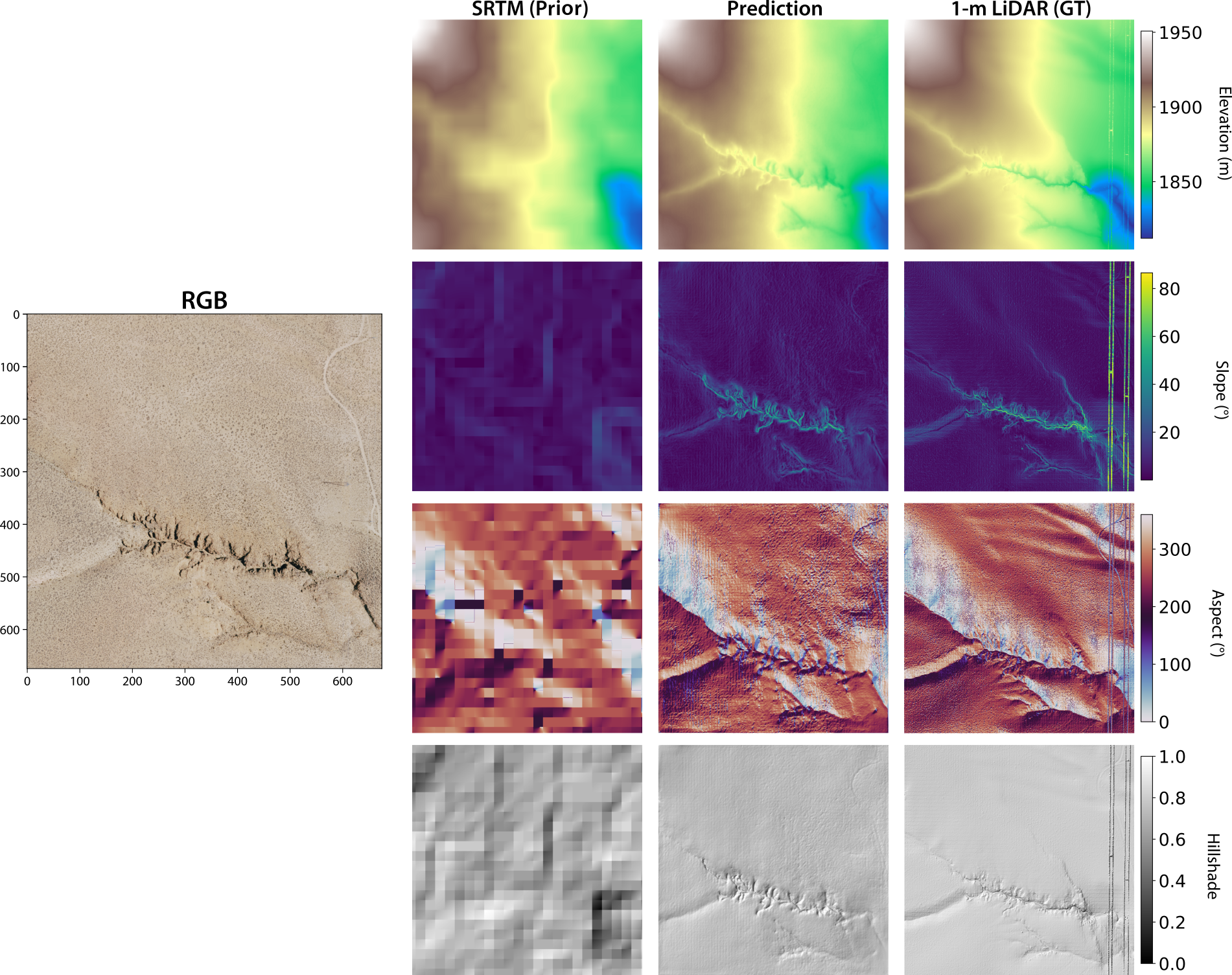}
    \caption{Visual Evaluation - DEM estimation in natural bare landscape, elevation in meters, hillslope gradient in degrees, aspect and radial direction, and hillshade visualization. The global SRTM general terrain was preserved for elevation prediction, while the RGB information added detailed terrain features, such as small streams which are clearly visible in the slope, aspect, and hillshade maps.}
    \textit{}
    \label{bare_low}
\end{figure}

Our suggested prior-based framework performance was evaluated by comparing absolute terrain feature errors between predicted DEMs, SRTM, and HR-LiDAR reference data. Overall, the predicted DEM matched the 0.6-m spatial resolution of the input RGB imagery, providing substantially greater topographic detail than the 30-m SRTM (Figures \ref{veg_low} and \ref{bare_low}).

\begin{table}[ht!]
\centering
\caption{Evaluation metrics for DEM estimation task across different landscape types. Predicted DEM and SRTM vs. LiDAR ground truth data across landscape types.}
\setlength{\tabcolsep}{1.0 pt}
\begin{tabular}{llcccc}
\hline
\textbf{} & \textbf{}  & \multicolumn{2}{c}{\textbf{Vegetated}} & \multicolumn{2}{c}{\textbf{Bare}} \\
\cline{3-4} \cline{5-6}
& & {Prediction} & {SRTM} & {Prediction} & {SRTM} \\
\hline
\multirow{2}{*}{\textbf{Elevation (m)}} 
& MAE   & \textbf{4.645} & 5.661 & \textbf{2.132} & 2.217 \\
& RMSE  & \textbf{5.928} & 7.275 & \textbf{2.695} & 2.871 \\
\hline
\multirow{2}{*}{\textbf{Slope (°)}} 
& MAE   & \textbf{17.762} & 19.033 & \textbf{2.721} & 2.860 \\
& RMSE  & \textbf{23.537} & 24.938 & \textbf{4.415} & 4.694 \\
\hline
\multirow{2}{*}{\textbf{Aspect (°)}} 
& MAE   & \textbf{81.039} & 83.065 & \textbf{60.269} & 63.505 \\
& RMSE  & \textbf{120.246} & 120.279 & \textbf{101.345} & 104.144 \\
\hline
\end{tabular}
\label{tab:gen_metrics_full}
\end{table}

\begin{figure*}[ht!]
    \centering
    \includegraphics[width=1\linewidth]{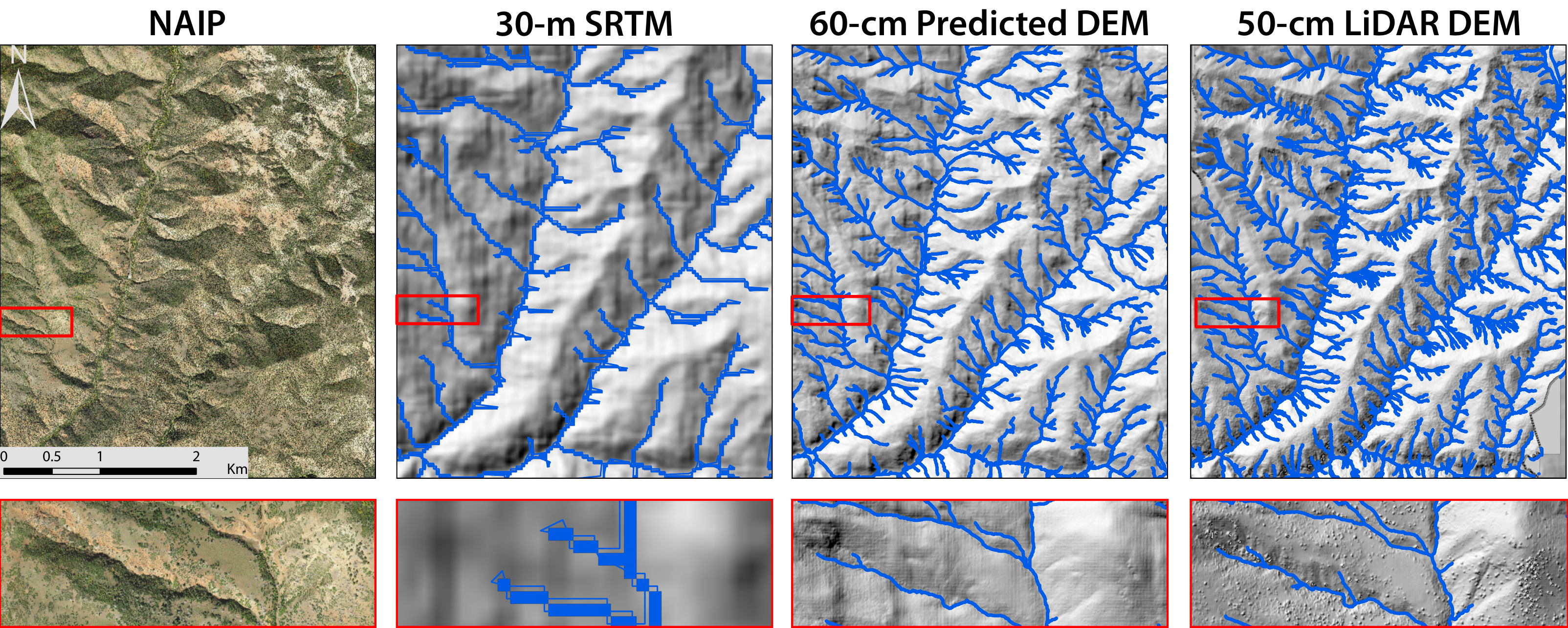}
    \caption{Seamless hillshaded DEM estimation showing the stream network calculated over the entire natural vegetated Diamond Fork Canyon site. Patch stitching did not affect the hydrological analysis, and the detail of the predicted DEM was a stream network that was visually comparable to that obtained with the 50-cm LiDAR DEM, as can be seen in the red zoomed-in visualization.}
    \textit{}
    \label{veg-streams}
\end{figure*}

\begin{figure*}[ht!]
    \centering
    \includegraphics[width=1\linewidth]{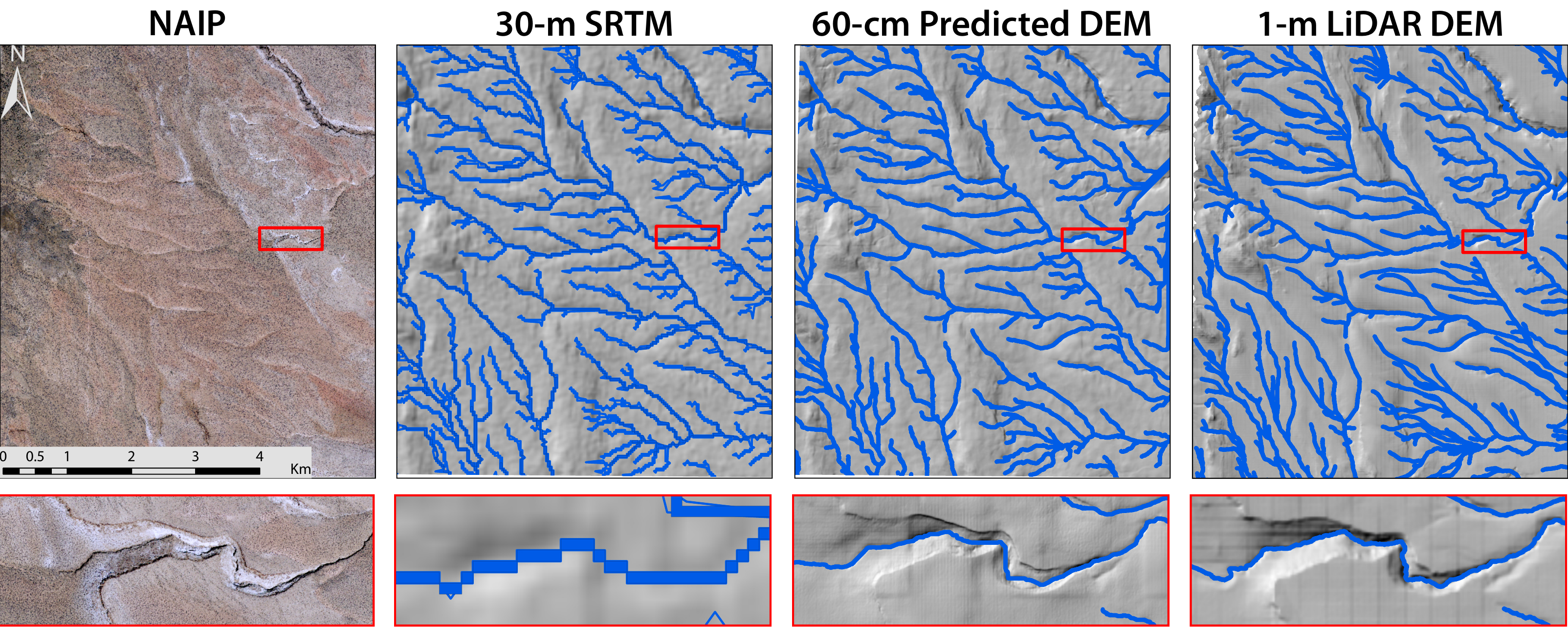}
    \caption{Visual Evaluation – Seamless hillshaded DEM estimation showing the stream network calculated over the entire natural bare terrain in the Casa Diablo Mountain site. Patch stitching did not affect the hydrological analysis, and the detail of the predicted DEM was a stream network visually comparable to the 1 m LiDAR DEM, as can be seen in the red zoomed-in visualization.}
    \textit{}
    \label{bare-streams}
\end{figure*}

\subsection{HR-DEM estimation}

Table \ref{tab:gen_metrics_full} shows the evaluation metrics across the two landscape types. The predicted DEM consistently outperformed SRTM across nearly all landscape types, as assessed against HR-LiDAR data.

In the vegetated landscape, the initial SRTM error was more significant, and therefore, the advantage of the predicted DEM was even more noticeable. This was especially true for  elevation and hillslope gradient estimations: the MAE dropped by almost 1 m from 5.661 m to 4.645 m, and the RMSE decreased from 7.275 m to 5.928 m. The MAE and RMSE of the hillslope gradient also improved, but the RMSE of the aspect was virtually the same.

For bare soil landscapes, the SRTM error was initially small in both cases of elevation, and the hillslope gradient errors were lowest in all types of landscape. Yet the accuracy of the predicted DEM was higher (elevation MAE: 2.132 m vs. 2.217 m; RMSE: 2.695 m vs. 2.871 m). The RMSE of the hillslope gradient improved slightly, and the aspect errors were markedly reduced (MAE: 60.269° vs. 63.505°; RMSE: 101.345° vs. 104.144°).

The metrics compared indicate that, in general, the prior-based framework substantially improved the absolute accuracy of the DEM, and this effect was more noticeable in the vegetated, although it also performed well in the simpler, bare soil landscapes (Figures \ref{veg_low} and \ref{bare_low}).

A comparative analysis of the different types of landscape showed that the accuracy of the elevation estimates was optimal in the bare soil landscapes. In the vegetated landscapes, however, elevation estimation was slightly less accurate relative to the DEM estimation benchmark. The improved accuracies of the hillslope gradient and aspect estimations in vegetated areas paralleled those observed in the primary DEM estimation task. See Figures \ref{veg_low}, and \ref{bare_low}.

\begin{table*}[ht!]
\centering
\caption{Accuracy assessment of the extracted stream networks at the natural vegetated evaluation site (Diamond Fork Canyon), using three buffer distances: 5 m, 10 m, and 25 m.}
\begin{tabular}{lcccccc}
\hline
\textbf{} & \multicolumn{2}{c}{{5 m}} & \multicolumn{2}{c}{{10 m}} & \multicolumn{2}{c}{{25 m}} \\
\cline{2-3} \cline{4-5} \cline{6-7}
& {Predicted DEM} & {SRTM} & {Predicted DEM} & {SRTM} & {Predicted DEM} & {SRTM} \\
\hline
IoU        & \textbf{0.360} & 0.127 & \textbf{0.472} & 0.227 & \textbf{0.627} & 0.378 \\
Precision  & \textbf{0.609} & 0.291 & \textbf{0.733} & 0.469 & \textbf{0.860} & 0.737 \\
Recall     & \textbf{0.468} & 0.183 & \textbf{0.571} & 0.306 & \textbf{0.699} & 0.437 \\
F1-score   & \textbf{0.529} & 0.225 & \textbf{0.642} & 0.370 & \textbf{0.771} & 0.549 \\
Accuracy   & \textbf{0.908} & 0.860 & \textbf{0.877} & 0.800 & \textbf{0.835} & 0.714 \\
\hline
\end{tabular}
\label{tab:streams_veg}
\end{table*}

\begin{table*}[ht!]
\centering
\caption{Accuracy assessment of the extracted stream networks at the natural bare evaluation site (Casa Diablo Mountain), using three buffer distances: 5 m, 10 m, and 25 m.}
\begin{tabular}{lcccccc}
\hline
\textbf{} & \multicolumn{2}{c}{{5 m}} & \multicolumn{2}{c}{{10 m}} & \multicolumn{2}{c}{{25 m}} \\
\cline{2-3} \cline{4-5} \cline{6-7}
& {Predicted DEM} & {SRTM} & {Predicted DEM} & {SRTM} & {Predicted DEM} & {SRTM} \\
\hline
IoU        & \textbf{0.218} & 0.118 & \textbf{0.308} & 0.218 & \textbf{0.470} & 0.422 \\
Precision  & \textbf{0.405} & 0.175 & \textbf{0.526} & 0.293 & \textbf{0.703} & 0.519 \\
Recall     & \textbf{0.320} & 0.268 & 0.427 & \textbf{0.460} & 0.587 & \textbf{0.692} \\
F1-score   & \textbf{0.357} & 0.212 & \textbf{0.471} & 0.358 & \textbf{0.640} & 0.593 \\
Accuracy   & \textbf{0.935} & 0.888 & \textbf{0.906} & 0.839 & \textbf{0.858} & 0.797 \\
\hline
\end{tabular}
\label{tab:streams_bare}
\end{table*}

\subsection{Surface and hydrological evaluation}

Stream network accuracy metrics for natural vegetated landscapes are presented in Table~\ref{tab:streams_veg}, while the corresponding metrics for bare landscapes are shown in Table~\ref{tab:streams_bare}. Seamless DEM reconstruction was achieved by unpatchifying predicted patches using a distance-based linear blending method. The resulting seamless DEMs for both sites are visualized in Figures~\ref{veg-streams} and \ref{bare-streams}. Stream networks were calculated using ArcGIS Pro and were compared with channel networks extracted from both SRTM and HR-LiDAR DEMs. Relative to the LiDAR data, the performance of the predicted DEM was consistently superior to that of the SRTM baseline data in both types of landscape.

According to the metrics in Table~\ref{tab:streams_veg}, the predicted DEM in vegetated landscapes provided a more accurate stream network compared with SRTM. This is evident from the IoU of extracted streams: the predicted DEM increased the IoU from 0.1265 to 0.3695. Similar improvements were observed in the terrain F1-score (0.2246 vs. 0.5289). As buffer size increased, all metrics improved noticeably. With a 25-m buffer, the predicted DEM achieved an IoU of 0.6274 compared with 0.3781 for SRTM, and an F1-score of 0.7711 compared with 0.5487 for SRTM. The same tendency was observed across all metrics, in which the accuracy of the predicted DEM consistently surpassed that of the SRTM-extracted stream network.

Likewise, according to the metrics in Table~\ref{tab:streams_bare}, the predicted DEM in bare soil landscapes also facilitated an extraction of a more accurate stream network compared with SRTM, as is evident from the IoU of extracted streams. The predicted DEM increased IoU from 0.1183 to 0.2176. Similar improvements are observed in the terrain F1-score (0.2116 vs. 0.3574). As the buffer size increased, all metrics improved. With a 25-m buffer, the predicted DEM achieved an IoU of 0.4703 compared with 0.4218 for SRTM, and an F1-score of 0.6397 compared to 0.5934 for SRTM. The same tendency of the predicted DEM consistently surpassing the stream network extracted from the SRTM was observed across all metrics.

Although MAE and RMSE of the natural vegetated landscapes were higher, the extracted stream networks in those landscapes were more accurate. Note that Diamond Fork Canyon is steeper, and compared with Casa Diablo Mountain, it is characterized by higher network density  (14 \,km/km² vs. 11 \,km/km²) and a higher mean slope  (59.73° vs. 32.23°).

Basin accuracy metrics for natural (bare) and vegetated landscapes are presented in Table~\ref{tab:basins_mean}. Overall, the predicted DEM consistently provides more accurate basin delineation compared with SRTM. In bare landscapes, both SRTM and the predicted DEM achieve relatively high accuracy; however, the predicted DEM outperforms SRTM, achieving an IoU of 0.9225 compared with 0.8700. 

\begin{figure*}[ht!]
    \centering
    \includegraphics[width=1\linewidth]{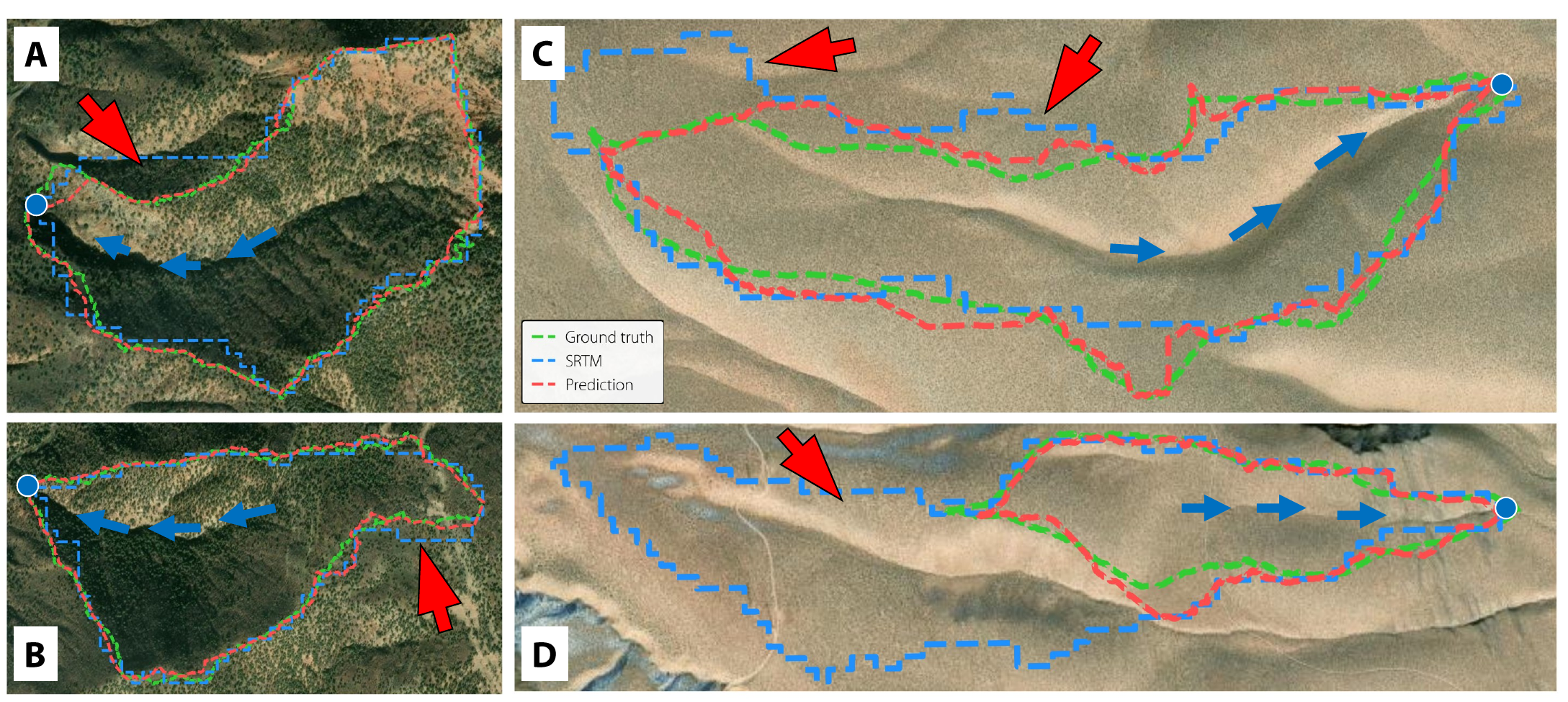}
    \caption{Comparison of sub-basin boundaries derived from predicted DEMs and SRTM against ground truth across four representative landscapes (A–D). Dashed contours indicate ground truth (green), SRTM-derived basins (blue), and predicted DEM basins (red). Red arrows highlight areas of notable disagreement, including boundary displacement and differences in contributing area extent. The results demonstrate improved alignment of the predicted DEM with ground truth compared to SRTM, particularly along complex terrain boundaries.}
    \textit{}
    \label{basins}
\end{figure*}

In vegetated landscapes, overall accuracy is lower for both datasets, reflecting an increased complexity of terrain representation under vegetation cover. Nevertheless, the improvement obtained using the predicted DEM is more pronounced, with an IoU of 0.8444 compared to 0.7301 for SRTM. In addition to the quantitative metrics, we provide four representative examples to visually illustrate the improvement in basin delineation (Fig.~\ref{basins}).

\begin{figure*}[ht!]
    \centering
    \includegraphics[width=1\linewidth]{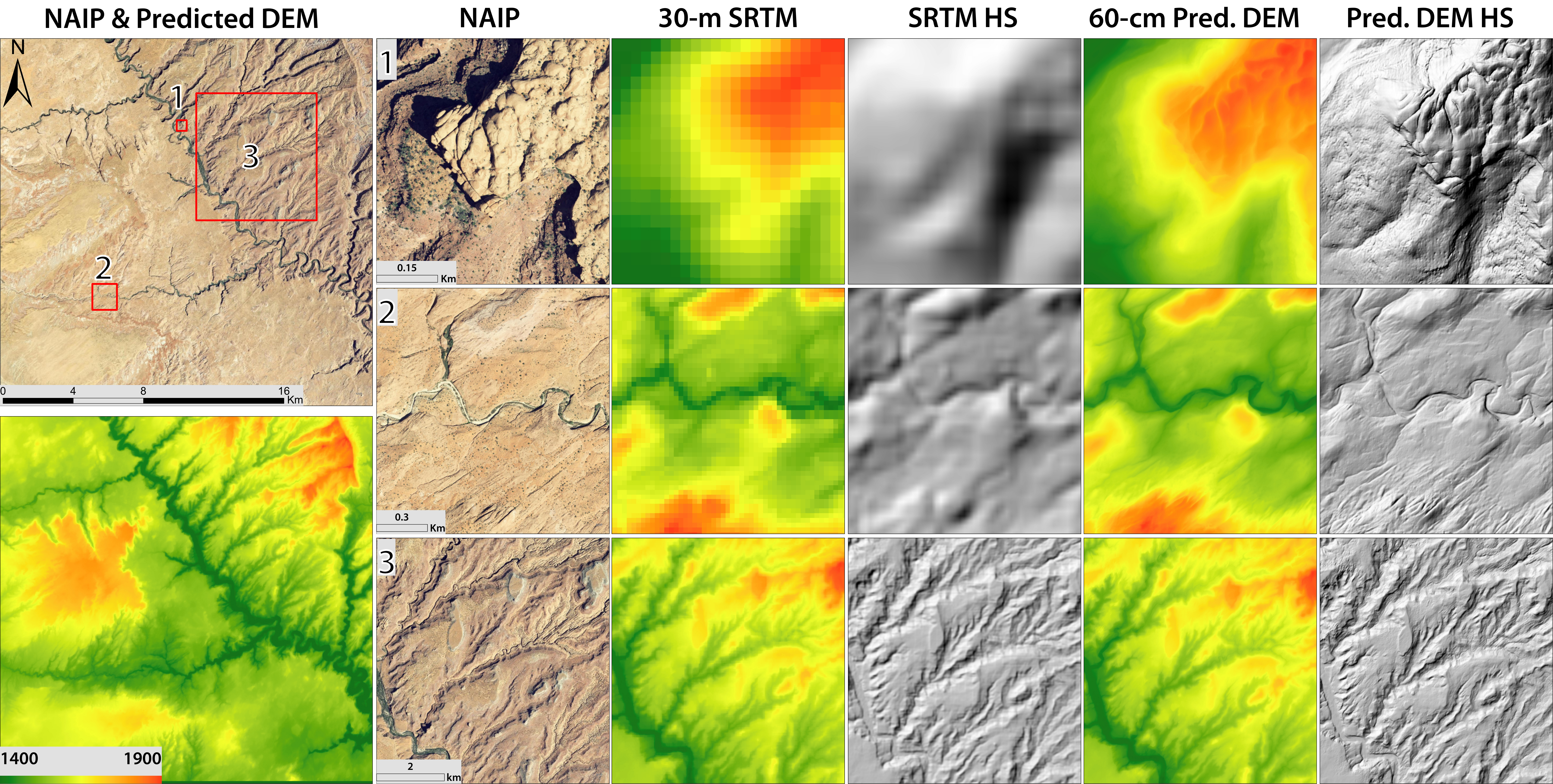}
    \caption{Visual Evaluation: Seamless hillshaded DEM estimation for the Escalante River region in Utah. Patch stitching is not visible, and the detail of the predicted DEM is visually comparable to that of the 30-m SRTM DEM, as can be seen in the red zoomed-in visualization.}
    \textit{}
    \label{Escalante_River}
\end{figure*}

\begin{figure*}[ht!]
    \centering
    \includegraphics[width=1\linewidth]{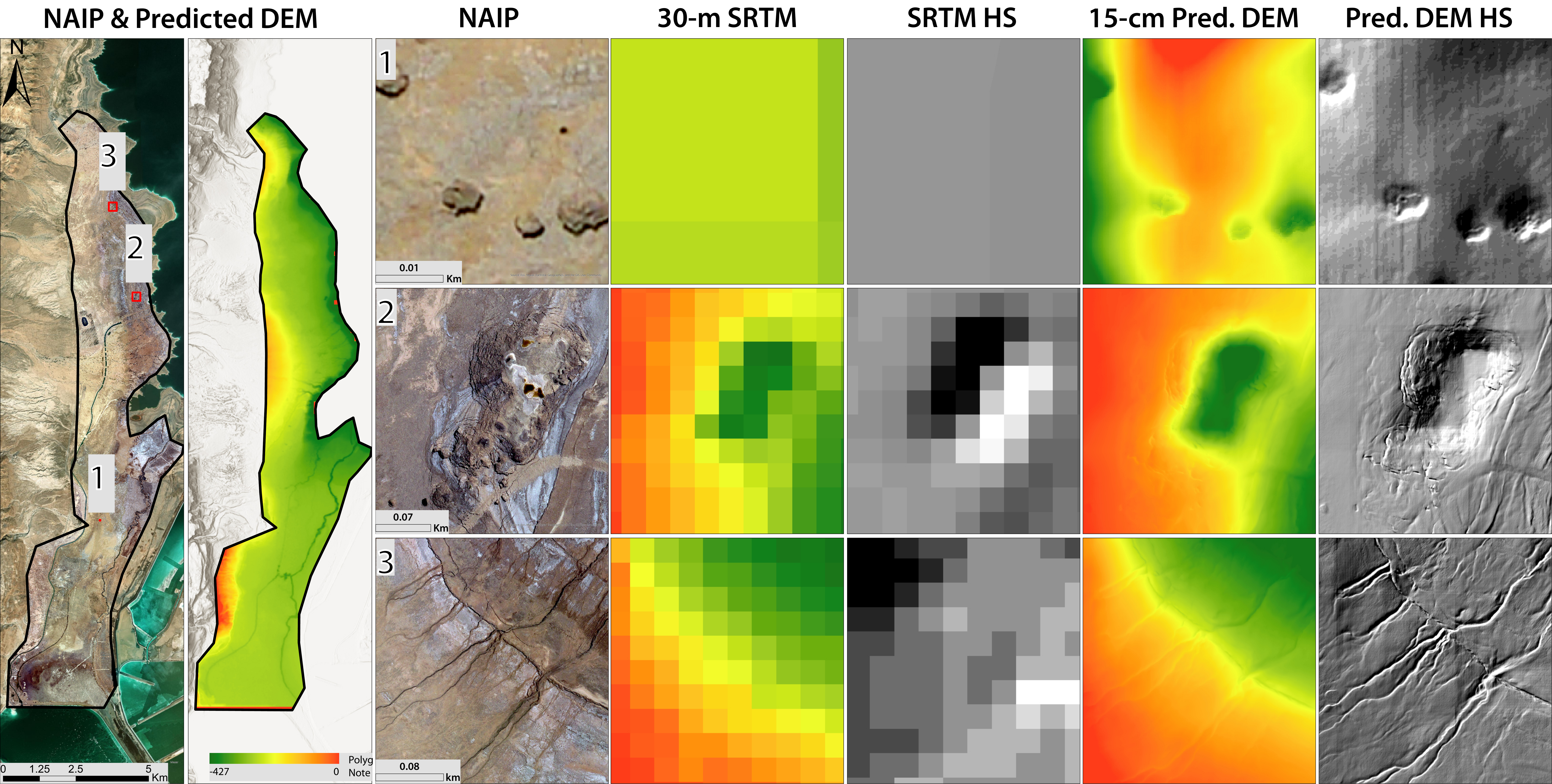}
    \caption{Visual Evaluation: Seamless hillshaded DEM estimation over the entire Dead Sea shore. Patch stitching is not visible, and the detail of the predicted DEM is visually comparable to that of the 30-m SRTM DEM, as can be seen in the red zoomed-in visualization.}
    \textit{}
    \label{deadsea_low}
\end{figure*}

Two large-scale seamless DEMs are visually presented in Figure~\ref{Escalante_River} for the Escalante River region in Utah and in Figure~\ref{deadsea_low} for the Dead Sea in Israel. DEMs quality is illustrated in the zoomed-in windows. The small streams and tributaries are clearly depicted in the 60-cm predicted DEM for the Escalante River and sinkholes and streams in 15-cm predicted DEM in the Dead Sea shore. The 30-m SRTM of both of these areas, however, did not capture such fine details. Moreover, tens of thousands of patches were successfully reconstructed into a seamless DEM. There is no visible evidence of patch boundaries. Global features, such as continuous stream networks that span many patches, are coherently preserved in the final DEM.

\subsection{Comparison with DEMSR methods}

Although our framework does not employ SR reconstruction, we compared our results to those obtained using the main DEM improvement approach, SR-DEM \citep{zhang2022comparison}. We identified a comparable dataset with reported results for this comparison. Current multi-modal SR methods typically use an upscaling factor of 4; consequently, the recently proposed MFSR model produces a fixed output resolution of 7.5\,m (upscaled from 30\,m), whereas our method achieves a finer resolution of 5\,m corresponding to the RGB resolution provided in the DEM-OPT-Depth SR dataset \citep{huang2025multimodal}. The values of RMSE as reported from various SR models on the DEM-OPT-Depth SR dataset are as follows: Bicubic – 14.891\,m, SRCNN – 8.346\,m, ESRGAN – 7.931\,m, TfaSR – 7.356\,m, SRFormer – 7.741\,m, and MFSR – 5.544\,m \citep{huang2025multimodal}. Using a publicly available subset of this dataset, our fine-tuned model achieved an RMSE of 5.535\,m using just 33\% of the data for fine-tuning, exceeding the reported results.

\begin{table}[ht!]
\centering
\caption{Accuracy assessment of the extracted basin extents (mean across all basins), for bare and vegetated sites.}
\setlength{\tabcolsep}{1.0 pt}
\begin{tabular}{lcccc}
\hline
\textbf{} & \multicolumn{2}{c}{\textbf{Bare}} & \multicolumn{2}{c}{\textbf{Vegetated}} \\
\cline{2-3} \cline{4-5}
& {Predicted DEM} & {SRTM} & {Predicted DEM} & {SRTM} \\
\hline
IoU        & \textbf{0.923} & 0.870 & \textbf{0.844} & 0.730 \\
Precision  & \textbf{0.962} & 0.940 & \textbf{0.903} & 0.835 \\
Recall     & \textbf{0.957} & 0.923 & \textbf{0.930} & 0.867 \\
F1-score   & \textbf{0.960} & 0.930 & \textbf{0.914} & 0.824 \\
\hline
\end{tabular}
\label{tab:basins_mean}
\end{table}

\section{Discussion}

\subsection{Performance in different landscape types}

\begin{figure}[ht!]
    \centering
    \includegraphics[width=1\linewidth]{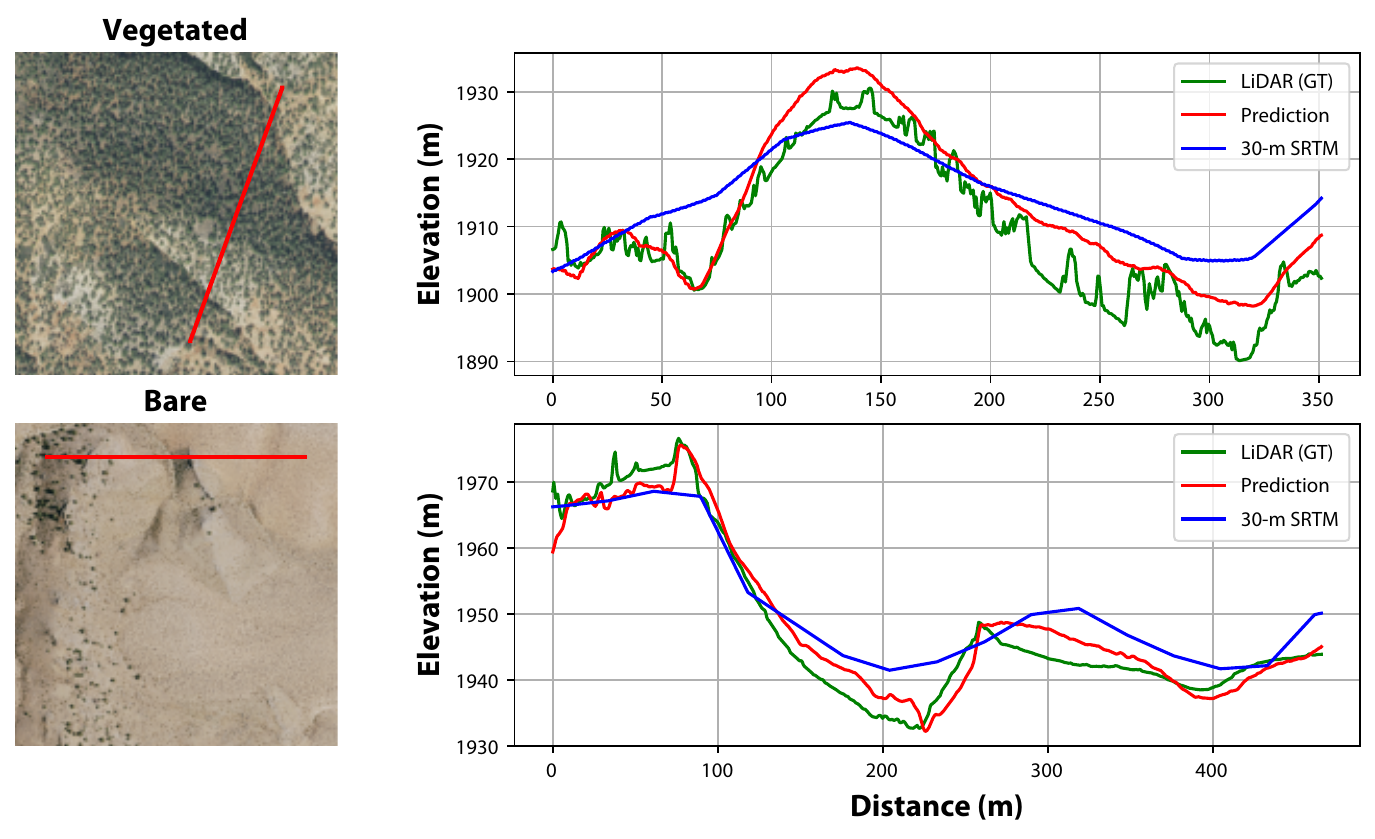}
    \caption{DEM estimation with SRTM, HR-LiDAR DEM, and predicted elevation profiles. Although SRTM missed fine features, the model estimation captured buildings and small streams in both the vegetated and bare terrain landscapes.}
    \textit{}
    \label{elevation_profiles}
\end{figure}

Our framework was evaluated across two landscape types with varying terrain complexity. The reliability of the global 30-m SRTM priors varied depending on terrain characteristics. Specifically, accuracy was lower in the more complex vegetated landscapes compared with bare terrain. This reduction was also observed in previous studies \citep{SU2014216, gamba2002srtm}. Specifically, SRTM vertical error was greater in regions containing natural or artificial features above the bare-earth surface \citep{hofton2006validation}. In vegetated landscapes, the recorded elevation values typically fall between ground surface elevation and that of the canopy top \citep{colosimo2009accuracy}. We observed similar trends. For instance, when compared with 50\,cm LiDAR DEM, SRTM had an RMSE of 7.275\,m in the vegetated Diamond Fork Canyon, whereas in the bare landscape of Casa Diablo Mountain, the RMSE was only 2.871\,m.

Although SRTM accuracy diminishes in complex terrains, MDE offers superior performance in capturing fine features, especially through cast shadow, self-shadow, or topographic shading \citep{Florea_2025, song2025enhancing, 10687819}. Thus, most significant improvements were observed in elevation and hillslope gradient accuracy for vegetated landscapes, with an 18\% improvement in MAE over SRTM. This is consistent with reported performance of MDE, which is more precise, particularly in delineating trees contours and in capturing local elevation changes \citep{hong2025depth2elevation, rs12172719}. Another observation supporting the superior performance of MDE in shaded areas is that the predicted stream network in the steep, vegetated landscape of Diamond Fork Canyon (see Fig.~\ref{veg-streams}) was more accurate than the predicted features in the flatter Casa Diablo Mountain site (see Fig.~\ref{bare-streams}). This suggests that, despite initial inaccuracies in the priors over vegetated landscapes, shadows enabled the model to infer more reliable topographic information, thereby improving local precision.

\subsection{MDE vs SR}

Super-resolution (SR) methods were widely employed to improve DEMs accuracy and spatial resolution in several studies \citep{s22030745}. SR approaches typically enhance DEM resolution through learned low-to-high resolution (LR-to-HR) translation. However, their performance is fundamentally constrained by the information contained in the low-resolution input and by predefined upscaling factors.

Recently, MDE has received increasing attention in RS. Despite its potential to reconstruct fine-scale terrain structure, MDE has rarely been applied in studies of open environments and has not yet been widely adopted for hydrological and geomorphological analyses \citep{10802125, RAFAELI20251}. One reason for this limited adoption is that MDE models generally operate in a local image context and therefore lack mechanisms to incorporate global elevation information \citep{Florea_2025}.

Several recent studies have attempted to bridge the gap between SR and MDE. For example, \citet{huang2025multimodal} proposed a multi-modal fusion super-resolution (MFSR) framework that integrates MDE into the SR process. Their approach utilizes predicted zero-shot relative depth to capture complementary structural terrain information, improving elevation reconstruction. Experimental results demonstrated a 24.63\% reduction in RMSE compared with state-of-the-art SR techniques. The MFSR framework represents an important step toward integrating monocular foundation models within DEM super-resolution workflows. However, the model remains constrained by a fixed 4$\times$ upscaling factor (i.e., from 30\,m to 7.5\,m resolution), which limits its ability to reconstruct finer terrain features and to generalize across RGB datasets \citep{huang2025multimodal}.

In contrast to DEMSR approaches, deep learning–based MDE models predict depth directly from a single RGB image using learned scene representations \citep{zhao2020monocular}. Rather than relying on LR-to-HR translation, these models exploit visual cues and contextual information within the RGB imagery to infer surface structure \citep{yang2024depth}. Consequently, MDE predictions are not restricted by predefined upscaling ratios and can produce depth estimates at the native resolution of the input RGB imagery (see Fig.~\ref{elevation_profiles}).

Within RS literature, MDE is often referred to as monocular height estimation (MHE). Such methods have been applied in agricultural and urban environments \citep{zhang2024towards}. For instance, canopy height estimation was demonstrated using a fine-tuned Depth Anything model, and more recently Depth Anything was applied to height estimation in urban RS scenes \citep{hong2025depth2elevation}. Despite these advances, MDE predictions still primarily rely on local visual context, which limits their ability to capture regional elevation trends across broader spatial scales \citep{song2025enhancing}.

A further challenge arises from the large spatial extent of RS imagery. Unlike ground-level applications where models process individual images, DEM prediction typically requires dividing large scenes into smaller patches during inference \citep{chen2025dynamicvis}. Without mechanisms to incorporate global elevation information, independently predicted patches may lead to inconsistencies in absolute elevation values across the reconstructed DEM \citep{hong2025depth2elevation}. Incorporating elevation priors therefore represents a promising strategy to preserve global elevation consistency while maintaining high-resolution terrain reconstruction.

Our results demonstrate that prior-based MDE can substantially overcome these limitations. By incorporating globally available elevation data (e.g., SRTM) as a prior, the proposed approach preserves absolute elevation context while allowing the model to exploit local visual depth cues. This enables seamless DEM reconstruction across patches and allows effective resolution enhancement by factors of approximately 50$\times$ to 200$\times$. Moreover, our method achieves lower RMSE values than SR approaches when evaluated on the DEM-OPT-Depth SR dataset.

\subsection{Large-scale application}

As noted earlier, high-resolution DEMs (HR-DEMs) are critical for many hydrological and environmental studies. In some cases, hazard assessments derived from available low-resolution DEMs (LR-DEMs) produced misleading results \citep{10.3389/feart.2015.00050}. LR-DEMs often contain substantial vertical errors and typically represent surface elevation—including vegetation and building structures—rather than bare-earth topography. This limitation can lead to systematic underestimation of flood risk in certain environments \citep{MUTHUSAMY2021126088}.

In the US, HR-DEMs are primarily provided through the USDA Natural Resources Conservation Service (NRCS) and the USGS 3D Elevation Program (3DEP) \citep{USGS_3DEP_lidar}. Nearly 80\% of the contiguous United States (CONUS) was mapped using LiDAR-derived DEMs at 1\,m resolution, and sub-meter DEMs are available in some regions \citep{USGS_3DEP_lidar}. Despite this extensive spatial coverage, these datasets have temporal limitations. Nationwide updates are not performed annually, which contrasts with the much higher acquisition frequency of RGB imagery from sources such as the National Agriculture Imagery Program (NAIP) or high-resolution satellite platforms \citep{sugarbaker2017status}.

Although the US benefits from extensive LiDAR coverage ($\leq 1\,\mathrm{m}$ resolution), comparable high-resolution DEM coverage is still lacking globally, primarily due to the high cost of airborne surveys \citep{ZANDSALIMI2025132687}. While such surveys are often impractical in low-resource regions, a variety of global or near-global DEM datasets have become freely available over the past two decades, alongside increasingly accessible high-resolution satellite imagery \citep{pu2012comparative}. Numerous studies have demonstrated that machine learning techniques can effectively refine LR-DEMs and correct systematic biases in these datasets \citep{YU2025104484, yue2024generative, tao2024simple, w12030816, 10500859}. Compared with LiDAR or photogrammetric acquisition methods, deep learning approaches that leverage these widely available datasets are significantly more resource-efficient and operationally flexible. Moreover, such models can be deployed on relatively modest computational resources, including personal workstations \citep{Ruiz-Lendínez31122023}.

Our results demonstrate that high-resolution, seamless DEMs can be generated at large spatial scales using the proposed framework. This capability was demonstrated in large-area applications. The method therefore has the potential to expand global HR-DEM coverage while also improving temporal resolution in regions where LiDAR data already exist but are updated infrequently. In addition to producing accurate and detailed elevation models, the framework is computationally efficient. On a single GPU, the system can generate DEMs at a rate of approximately 150\,km$^2$ per hour, and the use of high-performance computing clusters can further reduce processing time for continental-scale applications.



\section{Conclusions}

This study presents the tuning, evaluation, and large-scale application of a foundation model that leverages prior-based monocular depth estimation (MDE) to generate high-resolution, seamless, and metric DEMs across diverse landscapes. The proposed framework addresses three key tasks: high-resolution DEM estimation, void filling, and updating.

Extensive evaluation across three sites in the US representing both vegetated and bare-land environments indicates strong performance. Compared with SRTM, the proposed method reduced the mean absolute error (MAE) by 18\% in vegetated areas and by 3\% in bare-land regions. The approach also outperformed traditional super-resolution (SR) methods in both upscaling capability and overall accuracy. By combining MDE with globally contextualized patches, the framework enables seamless DEM reconstruction, making it suitable for downstream geospatial applications such as hydrological analysis.

Although we relied on publicly available datasets e.g., SRTM and NAIP imagery, both elevation priors and RGB inputs can be replaced with alternative data sources. For example, smaller upscaling factors could be explored, such as enhancing airborne LiDAR data using drone imagery. In addition, high-resolution satellite imagery could support large-scale mapping applications and further extend the use of deep learning–based MDE for global DEM generation. Future research may also include object-based evaluation using mapped features such as buildings and tree canopies to better characterize accuracy across different environments. Further improvements may arise from advances in model architectures, loss functions, and multimodal inputs, including the integration of multispectral data.

Overall, this work highlights the potential of MDE as a powerful alternative to traditional SR approaches for DEM enhancement. By bridging scene-aware visual representations with topographic information, the proposed framework opens new opportunities for generating prior-based, context-aware DEMs in heterogeneous and challenging terrains.

This proposed framework has important implications for large-scale hydrological applications that depend on terrain representation. High-resolution DEMs derived from this approach could improve drainage networks and watershed boundaries delineation, and extraction of topographic features that control runoff generation and flow routing across extensive regions. Such improvements may enhance regional flood hazard assessment, hydrological modeling, and landscape-scale analyses of surface water dynamics, particularly in areas where high-resolution elevation data are unavailable.

\section*{Acknowledgment}

We thank the Ministry of Agriculture, Chief Scientist Program, grant number 16-17-0005, 2022, and the Negev Scholarship from the Kreitman School of Ben-Gurion University of the Negev, for supporting Osher Rafaeli’s PhD studies.

\bibliographystyle{elsarticle-harv} 
\bibliography{cas-refs}




\end{document}